\pdfoutput=1

\documentclass[11pt]{article}

\usepackage{naacl2021}

\usepackage{times}
\usepackage{latexsym}

\usepackage[T1]{fontenc}

\usepackage[utf8]{inputenc}

\usepackage{microtype}

\usepackage{booktabs,chemformula}

\newcommand{\ours}{FiD-Ex}

\title{\ours: Improving Sequence-to-Sequence Models for Extractive Rationale Generation}

\author{
    Kushal Lakhotia$^\dagger\thanks{\hspace{.06in}Equal Contribution.}$\quad Bhargavi Paranjape$^\ddagger $\quad Asish Ghoshal$^\dagger$ \quad Wen-tau Yih$^\dagger$ \\  {\bf Yashar Mehdad}$^\dagger$\quad {\bf Srinivasan Iyer}$^\dagger{} ^*$ \\
    $^\dagger$Facebook AI \qquad $^\ddagger$University of Washington \\
    \texttt{\{kushall, aghoshal, scottyih, mehdad, sviyer\}@fb.com} \\ \texttt{bparan@cs.washington.edu} \\ 
 }

\begin{document}
\maketitle
\begin{abstract}
Natural language (NL) explanations of model predictions are gaining popularity as a means to understand and verify decisions made by large black-box pre-trained models, for NLP tasks such as Question Answering (QA) and Fact Verification. Recently, pre-trained sequence to sequence (seq2seq) models have proven to be very effective in jointly making predictions, as well as generating NL explanations. However, these models have many shortcomings; they can fabricate explanations even for incorrect predictions, they are difficult to adapt to long input documents, and their training requires a large amount of labeled data. In this paper, we develop \ours, which addresses these shortcomings for seq2seq models by: 1) introducing sentence markers to eliminate explanation fabrication by encouraging extractive generation, 2) using the fusion-in-decoder architecture to handle long input contexts, and 3) intermediate fine-tuning on re-structured open domain QA datasets to improve few-shot performance. \ours~significantly improves over prior work in terms of explanation metrics and task accuracy, on multiple tasks from the ERASER explainability benchmark, both in the fully supervised and in the few-shot settings.

\end{abstract}
\section{Introduction}

\begin{figure}[t]
    \centering
    \includegraphics[width=0.5\textwidth]{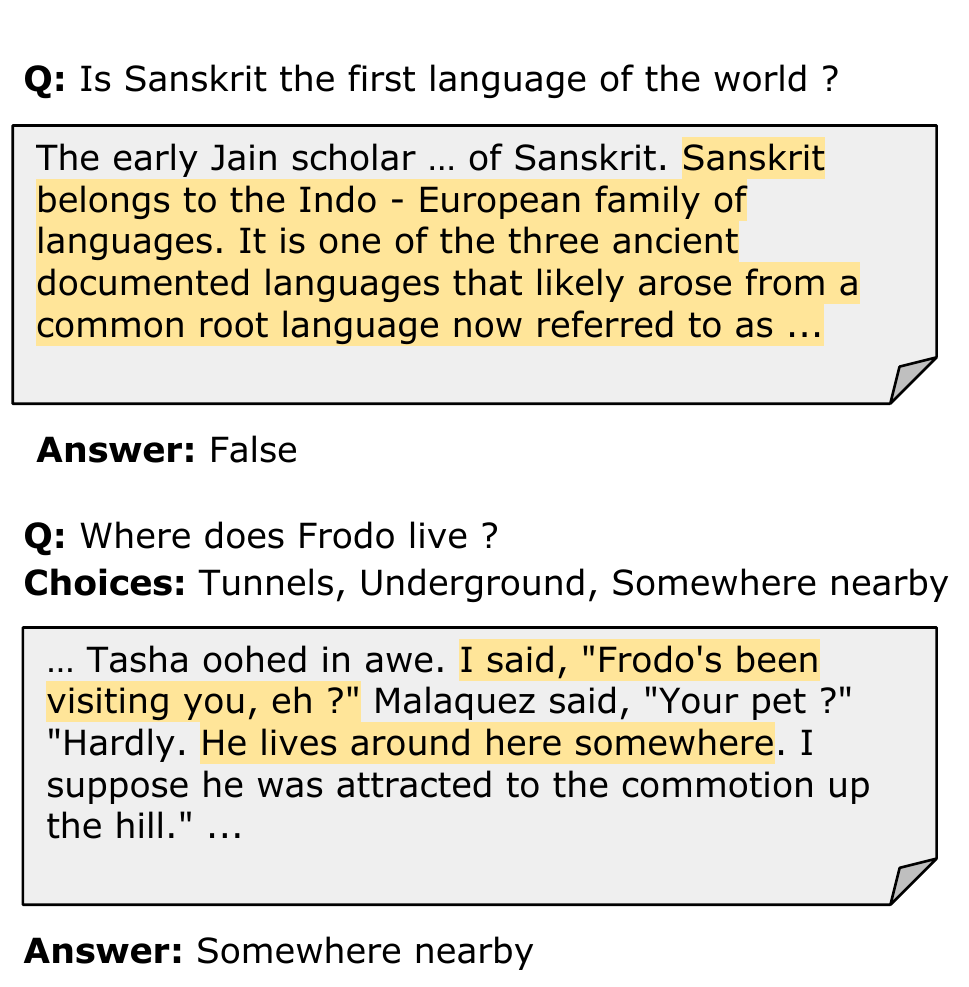}
    \caption{Example questions, answers, and corresponding passages from the BoolQ and MultiRC datasets from the ERASER benchmark \cite{deyoung-etal-2020-eraser}. Annotated rationales are highlighted. Note that rationales can be multi-sentence and non-contiguous.}
    \label{fig:examples}
\end{figure}

While large pre-trained language models \cite{devlin2019bert,raffel2019exploring,lewis-etal-2020-bart} with hundreds of millions of parameters have made super-human performance possible on various NLP datasets, they lack transparency into their decision making process, which can adversely affect user trust in their predictions. Recent works have proposed the use of textual rationales \cite{lei2016rationalizing,deyoung-etal-2020-eraser,latcinnik2020explaining} as a means to either obtain an understanding of the reasoning process of models, or as a human-readable snippet for users to verify predictions \cite{lipton2018mythos}. Figure \ref{fig:examples} presents examples of extractive textual rationales for two QA tasks from the ERASER dataset \cite{deyoung-etal-2020-eraser}\footnote{In this work, we use textual rationales and NL explanations interchangeably}. Recently, \newcite{narang2020wt5} show that sequence to sequence (seq2seq) models outperform previous methods at generating textual rationales for various explainability benchmarks. However, seq2seq models can fabricate rationales even for wrong predictions, are hard to scale to datasets involving several, long evidence documents, and, require large amounts of expensive rationale annotated data for training. In this paper, we introduce \ours, to alleviate these problems and enhance seq2seq models to achieve significant gains in rationale generation performance.

\newcite{camburu-etal-2020-make} find that models that generate free-form textual explanations can tailor them to convincingly justify incorrect model predictions, for example, generating ``There is no dog in the image'' to justify an \textit{no} prediction on the image of a dog. Although recent seq2seq NLP models \cite{narang2020wt5} obtain state of the art performance on rationale generation benchmarks, they are vulnerable to learning similar behaviours and can hallucinate new facts by tapping into stored world knowledge in the language model parameters. In order to retain their effectiveness and yet, alleviate the problem of explanation fabrication, \ours\ introduces the novel use of sentence markers into pre-trained seq2seq models. Training seq2seq models to decode sentence marker tokens in place of explanation tokens not only guarantees the production of unaltered rationales but also significantly improves explanation metrics on five datasets (see Section \ref{sec:results}).

Fine-tuning pre-trained models on data-rich intermediate tasks before fine-tuning on classification end tasks has recently been shown to improve end-task performance \cite{vu-etal-2020-exploring,pruksachatkun-etal-2020-intermediate}, more so in the few-shot setting. We find that this method also extends to seq2seq models, for explanation generation. We fine-tune pre-trained seq2seq  models to extract supporting evidence for existing reading comprehension datasets such as Natural Questions \cite{kwiatkowski-etal-2019-natural} and HotpotQA \cite{yang-etal-2018-hotpotqa}, which then improves downstream performance on rationale extraction benchmarks. This approach is motivated by the similarity of the process of gathering supporting facts for QA, to that of rationale extraction for classification tasks. While earlier works on rationale generation \cite{paranjape-etal-2020-information,narang2020wt5} are limited by the input passage size of pre-trained models and resort to input-passage truncation, \ours~uses the Fusion-in Decoder (FiD) approach \cite{izacard2020leveraging}, that separately encodes chunks of long input passages and fuses them in the decoder, which further improves performance. 

We combine these methods described above to develop \ours~({\bf Ex}tractive {\bf F}usion-{\bf i}n-{\bf D}ecoder). To summarize, \ours\ significantly improves upon the performance and trustworthiness of seq2seq models for rationale generation by 1) reducing their ability to fabricate explanations using sentence markers, 2) extending them to very long input passages, and, 3) intermediate fine-tuning on re-structured existing QA datasets. When applied to the ERASER datasets \cite{deyoung-etal-2020-eraser}, a popular benchmark for rationale extraction, \ours~yields significant gains on multiple tasks in terms of explanation metrics; an absolute token-F1 gain of 10\%  on Boolean Question Answering (BoolQ), 33.7\% on MovieReviews, 3.5\% on Evidence Inference, 2.6\% on FEVER, and 2.4\% on MultiRC, alongwith modest gains in terms of task accuracy, over prior work.

\begin{figure*}[ht!]
    \centering
    \includegraphics[width=\textwidth]{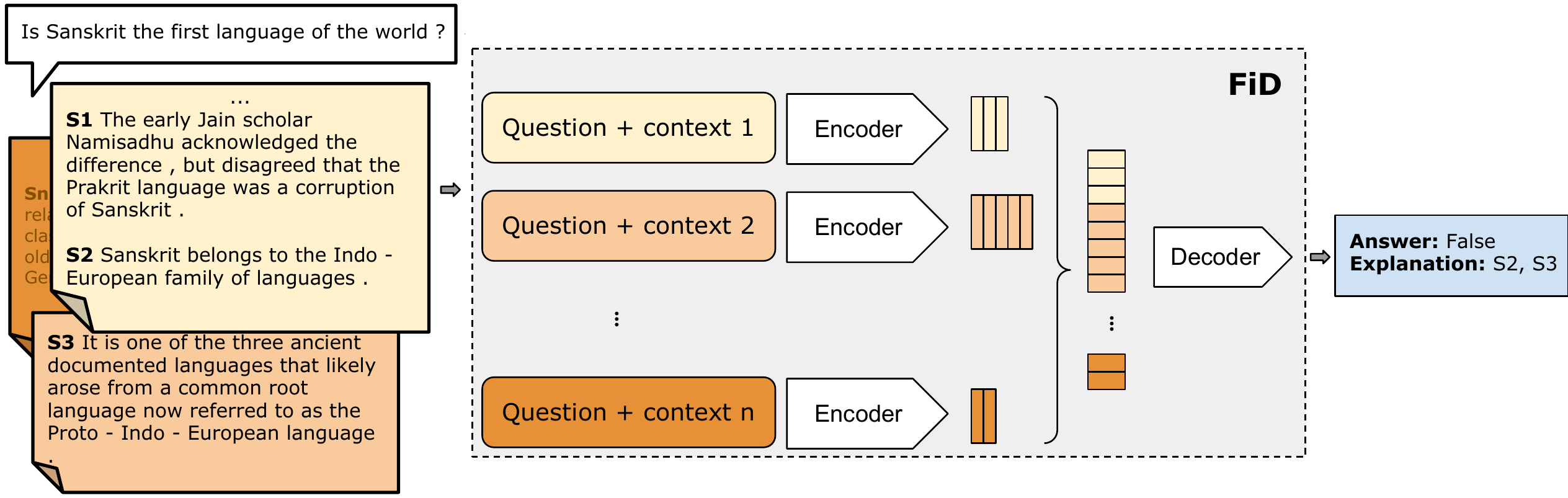}
    \caption {Fusion-in-Decoder architecture for rationale prediction. Each sentence from the passage is marked with sentence markers S1 ... SN. The passage is broken up into C contexts/chunks, which are passed to the encoder. The decoder then attends to the C concatenated and encoded passages to generate the output sequence. The output sequence is the classification token followed by rationale sentence markers. }
    \label{fig:model}
\end{figure*}

\section{Related Work}

Deep learning models typically function as black boxes offering very little insight into their decision making mechanics. To expose model understanding at various depths, researchers have proposed various structural probing \cite{tenney2018you,hewitt-manning-2019-structural,lin-etal-2019-open} and behavioral probing methods \cite{mccoy-etal-2020-berts,goldberg2019assessing,warstadt-etal-2019-investigating,ettinger-2020-bert}, as well as input saliency maps to highlight the most important tokens/sentences in the input for each prediction \cite{serrano-smith-2019-attention,ribeiro2016should,swanson2020rationalizing, tenney2019bert}, and input token relationships \cite{lamm2020qed}. Alongside, there is work on producing textual rationales \cite{lei2016rationalizing}, which are snippets of NL to help explain model predictions. Models may take a pipelined approach, where rationales are first selected as the sole inputs to the prediction stage, either in a supervised \cite{lehman2019inferring,pruthi2020weakly} or an unsupervised \cite{paranjape-etal-2020-information,bastings2019interpretable,jain2020learning} fashion. Alternatively, rationales can also serve as post-hoc supporting evidence, produced after the model prediction, as a snippet to help users verify the prediction \cite{yang-etal-2018-hotpotqa,thorne-etal-2018-fever}. In this work, we improve upon seq2seq models to produce the latter kind of NL explanations, along with model predictions.

In addition to extractive NL rationales obtained from subsequences of the input text, there is recent work on generating abstractive textual explanations for NLP tasks such as commonsense QA \cite{rajani2019explain} and NLI \cite{camburu2018snli,kumar2020nile}. \newcite{latcinnik2020explaining} train language models to transparently output their world knowledge as NL tokens, which is then consumed by a light-weight classifier. \citet{narang2020wt5} use a generative seq2seq T5 model to produce NL explanations token-by-token for the extractive ERASER benchmark, in order to take advantage of multi-task training i.e. training for prediction alone, as well as for explanations. Such models are susceptible to fabricating explanations to justify even their incorrect predictions, as identified by \newcite{camburu-etal-2020-make} and \newcite{wiegreffe2020measuring}. We introduce sentence markers into seq2seq models which alleviates this problem and also significantly improves their rationale extraction performance on sentence-level ERASER benchmark tasks.

Multiple prior works \cite{paranjape-etal-2020-information, jain2020learning, narang2020wt5} have explored methods to improve the performance of few-shot rationale generation, to reduce reliance on expensive rationale annotations. We fine-tune \ours~on re-structured intermediate open domain QA datasets to improve its regular and few-shot performance for rationale extraction. Fine-tuning large pre-trained models on intermediate tasks has been shown to be effective by prior work; \newcite{phang2018sentence} use data rich intermediate NLI tasks to improve target classification tasks; \newcite{talmor-berant-2019-multiqa} fine-tune on multiple QA datasets to improve the generalizability of QA models. Intermediate fine-tuning can also hurt performance \cite{bingel-sogaard-2017-identifying}.  \newcite{pruksachatkun-etal-2020-intermediate} recently present a large-scale study on fine-tuning a pre-trained RoBERTa model on 100 intermediate-target task combinations and use 25 probing tasks to understand the most desirable properties of intermediate tasks and datasets. \newcite{vu-etal-2020-exploring} explore transferability between 33 NLP tasks and produce task embeddings to help predict the most useful intermediate tasks for a target task. They conclude that intermediate tasks that require high levels of reasoning and inference abilities are more likely to help, particularly when task data is scarce. Closest to our method is \citet{kung2020zero}, who use Squad 2.0 as an intermediate learning task to fine-tune a shared encoder fitted with task-specific classification heads, for the downstream BeerReview and MovieReview rationalization tasks. Our approach is to strategically restructure large open domain QA datasets like Natural Questions and HotpotQA to make them amenable to intermediate fine-tuning of both the encoder and the decoder of pre-trained seq2seq models. This enables the use of exactly the same model architecture for multiple rationale prediction tasks.

\section{Modeling}
\label{sec:model}

In this section, we develop \ours, which improves upon the sequence to sequence approach to jointly produce NL rationales along with model predictions for tasks in the ERASER benchmark. Formally, given an input query $q$ and an input passage $p$ comprising sentences $p=\{s_j\}_{j=1}^{N}$, our goal is to produce a prediction $y$ and rationale sentences $\{e_k\}_{k=1}^{K}, e_k \in p, K << N$, that justify $y$. 

\newcite{narang2020wt5} fine-tune the pre-trained T5 (Text-to-Text Transfer Transformer) model \cite{raffel2019exploring} to auto-regressively produce the prediction and the explanation in a token-by-token fashion. Specifically, their model takes an input of the form ``explain \{task-name\}: $q$ $p$'', represented as a sequence of subword units \cite{sennrich-etal-2016-neural} using SentencePiece \cite{kudo-richardson-2018-sentencepiece}, and is trained to auto-regressively maximize the likelihood of an output sequence represented as ``\{prediction\} explanation: $e_1 \cdots$ explanation: $e_K$''. An input from the BoolQ dataset \cite{clark-etal-2019-boolq} might be represented as ``explain boolq: Is Sanskrit the first language of the world? <passage-tokens>'', with the output represented as ``False explanation: Sanskrit belongs to the Indo-European family of languages. explanation: Is is one of the three ...'' Such a model can be trained on data, both with and without explanation annotations, by dropping the unavailable parts of the output sequence. This model achieves state of the art explanation performance on several ERASER tasks and serves as a strong baseline which we build upon.

\subsection{Fusion-in-Decoder Approach} 
\label{sec:fid}
Current approaches typically truncate $p$ to 512 or 1024 tokens, which is particularly limiting for passages from datasets such as BoolQ, that use very long input passages (> 3000 tokens). To accommodate longer input passages, both for intermediate fine-tuning (see Section \ref{sec:ift}) and target fine-tuning, we use the Fusion-in-Decoder (FiD) architecture of \newcite{izacard2020leveraging}  as a replacement for the single encoder-decoder model of \newcite{narang2020wt5}. Using FiD, we break $p$ into smaller chunks and encode each chunk independently using the pre-trained T5 encoder (see Figure \ref{fig:model}). This expands the effective input length of the encoder, and at the same time, keeps computation resources growing linearly with the number of passages as opposed to quadratically. These separately encoded representations are then fused in the decoder, which then attends to all passage tokens, when producing output tokens. For encoding, we concatenate the query $q$ with each chunk of the input passage $p$. Further, we also prefix query and context tokens with special tokens, "question:" and "passage:" respectively. Making use of additional context from the passage, without truncation, significantly improves performance on the intermediate fine-tuning tasks as well as on the BoolQ, Movie Reviews and Evidence Inference end tasks (see Table \ref{tab:results}).

\subsection{Sentence Markers} \newcite{narang2020wt5}, as well as other works \cite{camburu-etal-2020-make}, point out that seq2seq models can fabricate reasonable sounding rationales to justify their incorrect predictions. To alleviate this issue, we introduce sentence markers into the input and output to enable the model to learn to generate a rationale sentence as a single unit. This technique has the added benefit that the rationales produced by the model are guaranteed to be strictly extractive at the sentence level, while retaining the performance benefits of a seq2seq architecture. Specifically, we preprocess the input passage $p$ by prefixing each sentence $s_i$ with a sentence marker token \texttt{S\{i\}}. Sentence markers are added to the passage before subdividing the passage into multiple chunks for encoding (as in Section \ref{sec:fid}). We also train the decoder to output the special sentence marker tokens, instead of NL tokens. Thus, the input is represented as ``question: $q$ passage: S1 $s_1$ S2 $s_2 \cdots $ SN $s_N$'' and the output as ``False explanation: $S_{e_1} \cdots$ explanation: $S_{e_K}$''. The example from BoolQ would be represented as ``explain boolq question: Is Sanskrit the first language of the world passage: S1 <Sent-1> ... SN <Sent-N>'' and the output as ``False explanation: S2 explanation: S3''. Note that these markers are injected as NL text, and would be later split into sub-word units. During inference, sentence markers are produced and mapped back to the corresponding sentences from the input.

\subsection{Intermediate Fine-tuning} 
\label{sec:ift}
Since obtaining rationale annotations for datasets is expensive, we look to fine-tune on existing large datasets to improve target task performance, particularly in the few-shot setting. Specifically, we re-structure open-domain QA datasets with answer span annotations to follow the same input-output structure as our target tasks, i.e., we produce a dataset of (query $q$, passage $p$, prediction $y$, and extractive rationales $e$) tuples from existing ODQA datasets. The datasets, together with their specific re-structuring methods, are described in Section \ref{sec:datasets}. In our experiments, we first fine-tune \ours\ on a combination of multiple ODQA datasets, and finally, fine-tune on our target evaluation task.

\section{Datasets} \label{sec:datasets}

In this section, we discuss the open-domain QA datasets and our pre-processing steps to prepare them for intermediate fine-tuning, as well as, the ERASER rationalizing datasets that we use for evaluation. Table \ref{tab:dataset_stats} presents the number of examples in each dataset split, as well as the average input passage lengths, in terms of the number of tokens and sentences, for both types of datasets.

\begin{table}
  \small
  \begin{tabular}{lrrrr}
    \toprule
    Dataset & Train & Val & Test & Toks / Sents \\
    \midrule
    NQ & $69,662$ & $4,352$ & - & $1,782$ / $66$ \\
    HotpotQA & $180,894$ & $14,810$ & - & $1,649$ / $75$ \\
    \midrule
    BoolQ & $6,363$ & $1,491$ & $2,807$ & $3,391$ / $165$ \\
    Movies & $1,600$ & $200$ & $200$ & $774$ / $37$ \\
    EVI & $7,958$ & $972$ & $959$ & $4,658$ / $153$ \\
    MultiRC & $24,029$ & $3,214$ & $4,848$ & $300$ / $14$ \\
    FEVER & $97,957$ & $6,122$ & $6,111$ & $288$ / $11$ \\
    \bottomrule
  \end{tabular}
  \caption{Dataset Statistics: Overview of dataset sizes in terms of the number of examples in train, validation and test sets, for our intermediate fine-tuning (top) datasets and evaluation (bottom) datasets. We also compare the respective passage lengths in terms of number of input tokens and sentences.}
  \label{tab:dataset_stats}
\end{table}

\subsection{Intermediate Fine-Tuning Datasets}

\paragraph{Natural Questions (NQ)} \cite{kwiatkowski-etal-2019-natural} is a dataset of real Google search queries with answer-span annotations from Wikipedia pages. Following \newcite{lee-etal-2019-latent} we use a subset containing short answers (at most 5 tokens). For every question and answer-span annotation, we use the question as $q$, the segmented Wikipedia passage as $p$, the answer tokens as the prediction $y$, and the single sentence containing the answer span as the rationale $e$. The Wikipedia passage is processed to remove all tables and lists, while we retain section headers.
 
\paragraph{HotpotQA} \cite{yang-etal-2018-hotpotqa} is a multi-hop QA dataset, where each question and answer annotation is accompanied with supporting fact sentence annotations from multiple Wikipedia documents. Similar to NQ, we use the question as $q$ and the answer tokens as the prediction $y$. However, since there are multiple Wikipedia pages with evidence, we treat each Wikipedia page as a separate passage $p$ and aggregate the annotated rationale sentences from that page as the rationales $e$. Thus, a single HotpotQA (question,answer) tuple produces as many examples as Wikipedia pages that are part of its supporting facts.

\subsection{Evaluation Data}

We evaluate on a subset of the datasets released as part of the ERASER benchmark \cite{deyoung-etal-2020-eraser} which comprise an input query and passage, an output class label, and input sentences annotated as rationales. We discuss these datasets in this section.

\paragraph{BoolQ} \cite{clark-etal-2019-boolq} comprises questions, whose answers can be either True or False, paired with long Wikipedia passages (> 3000 tokens), as well as sentence level rationale annotations (provided by ERASER) that support the answer.

\paragraph{MultiRC} \cite{khashabi2018looking} is a dataset of questions with multiple-choice answers, and an input passage annotated with sentence level rationales. This dataset is evaluated as a Boolean QA task by concatenating each answer choice to the question $q$, and assigning a True label to the correct choices and False to the others. All instances use the same set of supporting facts independent of the answer.
 
\paragraph{MovieReviews (Movies)} \cite{zaidan-eisner-2008-modeling,pang2004} contains movie reviews paired with binary positive/negative labels, without a query $q$ (we set it to ``What is the sentiment of this review?'' in our models). While ERASER provides span-level rationale annotations, we translate these to sentence level annotations following prior work \cite{paranjape-etal-2020-information}.

\paragraph{FEVER} \cite{thorne-etal-2018-fever} The ERASER version of the FEVER dataset contains input passages together with claims ($q$) that must be classified as supported or refuted, based on the passage, together with sentence level rationale annotations from the input passage.
 
\paragraph{Evidence Inference (EVI)} \cite{lehman2019inferring}  comprises (intervention, outcome, comparator) triples (concatenated as $q$) together with randomized controlled trial articles (> 4000 tokens), with the predictions being whether the intervention significantly increases, decreases, or has no effect on the outcome with respect to the comparator of interest. ERASER provides sentence level supporting facts on a subset of this dataset. 

We do not evaluate on the ERASER datasets of \textbf{e-SNLI} and \textbf{CoS-E} since they only use single sentence input passages.

\section{Evaluation Metrics}
\label{sec:metrics}

For all datasets, we report Exact Match Accuracy (EM) in terms of exact token match between the predicted class label and the true label, which corresponds directly to classification accuracy on the original task. To evaluate the quality of explanations, we report the following: 

\paragraph{Rationale F1 (RF1)} is an F1 score computed on the set of predicted explanation sentences as compared to the set of gold explanation sentences, computing set intersection based on exact sentence match.

\paragraph{Token F1 (TF1)} is an F1 score computed at the token level between the predicted explanation sentence tokens and the gold explanation sentence tokens, in terms of sets of token positions, by first mapping tokens to token positions in the input passage. When using sentence markers, we map the markers back to the original sentences before computing TF1. The sentences are tokenized using the \texttt{spaCy} tokenizer following \newcite{narang2020wt5}.

\paragraph{Intersection over Union (IOU F1)} as described in \newcite{deyoung-etal-2020-eraser} is computed by first matching up each predicted rationale with a gold rationale, and then computing F1. This is similar to Rationale F1, except that the match function is not exact match. A prediction and gold sentence match if the size of the overlap of their token positions divided by the size of the union of the token positions is higher than a threshold (we use 0.5). For our models, IOU F1 is very similar in magnitude to Rationale F1.
\begin{table} [ht!]
  \small
  \begin{tabular}{lcccc}
    \toprule
      & EM & RF1 & IOU F1 & TF1 \\
    \midrule
    \textbf{BoolQ} & $$ & $$ & $$ & $$ \\
    {C=1, No SM} & $65.4$ & $42.9$ & $45.9$ & $46.9$ \\
    {C=1, With SM} & $72.9$ & $49.7$ & $49.7$ & $50.3$ \\
    {C=10, With SM} & $73.7$ & $51.5$ & $51.6$ & $52.5$  \\
    {+ IFT} & $\bf76.3$ & $\bf56.6$ & $\bf56.6$ & $\bf57.1$ \\
    \addlinespace[0.5ex]
    \cline{2-5}
    \addlinespace[0.5ex]
    {C=10, With SM, 25\%} & $68.9$ & $50.7$ & $50.8$ & $51.7$ \\
    {+ IFT} & $\bf73.9$ & $\bf54.2$ & $\bf54.3$ & $\bf54.9$ \\
    \midrule
    \textbf{Movie Reviews} & $$ & $$ & $$ & $$  \\
    {C=1, No SM} & $89.5$ & $20.1$ & $27.2$ & $30.0$ \\
    {C=1, With SM} & $89.0$ & $55.7$ & $55.9$ & $57.7$ \\
    {C=6, With SM} & $97.0$ & $64.4$ & $64.5$ & $\bf66.7$ \\
    {+ IFT} & $\bf97.5$ & $\bf64.7$ & $\bf64.7$ & $66.4$ \\
    \addlinespace[0.5ex]
    \cline{2-5}
    \addlinespace[0.5ex]
    {C=6, With SM, 25\%} & $\bf50$ & $\bf48.5$ & $\bf48.5$ & $\bf50.8$ \\
    {+ IFT} & $50$ & $48.4$ & $48.4$ & $50.6$ \\
    \midrule
    \textbf{Evidence Inference} & $$ & $$ & $$ & $$  \\
    {C=1, No SM} & $64.2$ & $14.9$ & $15.2$ & $14.8$ \\
    {C=1, With SM} & $63.3$ & $29.6$ & $29.6$ & $29.4$ \\
    {C=10, With SM} & $74.1$ & $49.9$ & $49.9$ & $49.8$ \\
    {+ IFT} & $\bf75.3$ & $\bf50.6$ & $\bf50.6$ & $\bf50.5$ \\
    \addlinespace[0.5ex]
    \cline{2-5}
    \addlinespace[0.5ex]
    {C=10, With SM, 25\%} & $72.5$ & $47.7$ & $47.8$ & $47.8$ \\
    {+ IFT} & $\bf72.5$ & $\bf49.3$ & $\bf49.4$ & $\bf49.5$ \\
    \midrule
    \textbf{MultiRC} & $$ & $$ & $$ & $$  \\
    {C=1, No SM} & $78.3$ & $66.6$ & $67.5$ & $67.2$ \\
    {C=1, With SM} & $78.7$ & $\bf72.6$ & $\bf72.6$ & $\bf72.3$ \\
    {+ IFT} & $\bf79.8$ & $72.2$ & $72.2$ & $71.8$ \\
    \addlinespace[0.5ex]
    \cline{2-5}
    \addlinespace[0.5ex]
    {C=1, With SM, 2k} & $77.4$ & $66.4$ & $66.4$ & $65.9$ \\
    {+ IFT} & $\bf76.5$ & $\bf68.8$ & $\bf68.8$ & $\bf68.8$ \\
    \midrule
    \textbf{FEVER} & $$ & $$ & $$ & $$  \\
    {C=1, No SM} & $\bf92.8$ & $70.1$ & $71.2$ & $71.0$ \\
    {C=1, With SM} & $92.7$ & $83.4$ & $83.4$ & $83.3$ \\
    {+ IFT} & $92.5$ & $\bf83.8$ & $\bf83.8$ & $\bf83.8$ \\
    \addlinespace[0.5ex]
    \cline{2-5}
    \addlinespace[0.5ex]
    {C=1, With SM, 2k} & $89.0$ & $80.1$ & $80.1$ & $80.0$ \\
    {+ IFT} & $\bf89.6$ & $\bf81.2$ & $\bf81.2$ & $\bf81.1$ \\
    \bottomrule
  \end{tabular}
  \caption{Performance of our models. We compare the effects of using a larger context (C), sentence markers (SM), and intermediate fine-tuning (IFT).}
  \label{tab:results}
\end{table}
\begin{table*} [ht!]
    \centering
    \small
    \label{tab:standard}
    \begin{tabular}{l|ccc|ccc|ccc}
    \toprule
        \multicolumn{1}{c}{\bf Model} & \multicolumn{3}{c}{\bf BoolQ} & \multicolumn{3}{c}{\bf Movie Reviews} & \multicolumn{3}{c}{\bf Evidence Inference}  \\
        \cmidrule(l{3pt}r{3pt}){1-10}
        & EM & IOU F1 & Token F1 & EM & IOU F1 & Token F1 & EM & IOU F1 & Token F1 \\
        Bert-to-Bert & $54.4$ & $5.2$ & $13.4$ & $86.0$ & $7.5$ & $14.5$ & $70.8$ & $45.5$ & $46.8$ \\
        WT5 Base & $-$ & $-$ & $-$ & $\bf98.0$ & $-$ & $32.7$ & $-$ & $-$ & $-$ \\
        IB Supervised & $63.4$ & $32.3$ & $19.2$ & $85.4$ & $43.4$ & $28.2$ & $46.7$ & $13.3$ & $10.8$ \\
        \ours\ Base & $\bf76.3$ & $\bf56.6$ & $\bf57.1$ & $97.5$ & $\bf64.7$ & $\bf66.4$ & $\bf75.3$ & $\bf50.6$ & $\bf50.5$ \\
        \bottomrule
    \end{tabular}
\end{table*}
\begin{table*}[ht!]
    \centering
    \small
    \label{tab:standard}
    \begin{tabular}{l|ccc|ccc}
    \toprule
        \multicolumn{1}{c}{\bf Model} & \multicolumn{3}{c}{\bf MultiRC} & \multicolumn{3}{c}{\bf FEVER} \\
        \cmidrule(l{3pt}r{3pt}){1-7}
        & EM & IOU F1 & Token F1 & EM & IOU F1 & Token F1 \\
        Bert-to-Bert & $63.3$ & $41.6$ & $41.2$ & $87.7$ & $83.5$ & $81.2$ \\
        WT5-Base & $77.8$ & $-$ & $69.9$ & $-$ & $-$ & $-$ \\
        IB Supervised & $66.4$ & $54.4$ & $54.0$ & $88.8$ & $66.6$ & $63.9$ \\
        \ours\ & $\bf78.7$ & $\bf72.6$ & $\bf72.3$ & $\bf92.5$ & $\bf83.8$ & $\bf83.8$ \\
        \bottomrule
    \end{tabular}
    \caption{Performance of our best \ours\ model (multi-context input, sentence markers, and intermediate fine-tuning) compared with prior work. For WT5, we use their base model since we report all our metrics using T5-base. IOU F1 for IB is reported using a threshold of 0.1 whereas we report all our IOU metrics using a stricter threshold of 0.5.}
    \label{tab:comparison}
\end{table*}
\section{Implementation Details}

We use the FiD \cite{izacard2020leveraging} model architecture with the base version of T5. We use $512$ input sub-word tokens per context for BoolQ, Movie Reviews, Evidence Inference and FEVER, and $1024$ for MultiRC. For increasing passage size, we use context counts of $10$ for BoolQ and Evidence Inference, and $6$ for Movie Reviews. We do not see significant benefits from using increased passage size for MultiRC and FEVER. We run all experiments on machines with $8$ 32-GB GPUs using data distributed training with a batch size of $8$ per gpu. 
We train all models for a total of 20000 steps using the Adam \cite{kingma2014adam} optimizer. We choose learning rates from $\{1e^{-4}, 1e^{-5}\}$ based on validation performance and use linear decay. We compute validation set metrics every 500 steps and select the model with the best validation Token-F1 score. We use greedy decoding to decode the prediction and the explanation in our FiD decoder. The above settings are used, both for intermediate fine-tuning as well as for end-task fine-tuning. For segmenting Wikipedia passages into sentences for NQ, we use the pre-trained Punkt \cite{kiss-strunk-2006-unsupervised} sentence segmenter for English from the \texttt{nltk} library. For our evaluation datasets, we used the pre-segmented and pre-tokenized input passages provided by ERASER.

\section{Results and Discussion}
\label{sec:results}

We compare the performance of different variants of our \ours\ model using all evaluation metrics on five ERASER datasets, in Table \ref{tab:results}. The first model for each dataset can be viewed as our re-implementation of the model of \newcite{narang2020wt5} i.e., T5 with a single context, and without sentence markers. We describe all results in this section in terms of Token-F1 (TF1) but similar trends are observed for the other rationale metrics. We report all gains in absolute percentage points.

\paragraph{Sentence Markers} The addition of sentence markers leads to a significant improvement in explanation metrics on all five datasets as compared to generating raw tokens; BoolQ TF1 improves by 3.4\%, Movie Reviews by 27.7\%, Evidence Inference by 14.6\%, MultiRC by 5.1\% and FEVER by 12.3\%. Additionally, it also provides the desirable guarantee of being extractive by eliminating the problem of fabricated rationales that seq2seq models are susceptible to.

\paragraph{Increased Passage Size} By using multiple context encoders enabled by the FiD architecture, as opposed to truncation methods used by prior work, we are able to significantly improve performance; BoolQ TF1 improves by 2.2\%, Movie Reviews by 9\% and Evidence Inference by 20.4\%. This is accompanied by task EM improvements of 8\% in Movie Reviews and 10.8\% in Evidence Inference. Input passages in MultiRC and FEVER are not long enough to benefit significantly from increased passage size.

\paragraph{Intermediate Fine-tuning and Few-shot Performance} We perform intermediate fine-tuning (IFT) on a combined dataset of Natural Questions and HotpotQA that has been re-formatted for rationale extraction tasks. Note that IFT is also done using sentence markers. Fine-tuning on the full training sets of our evaluation tasks improves rationale metrics by 4.6\% for BoolQ. A more marked improvement is seen in the few-shot setting, where we fine-tune using only 25\% data for the BoolQ and Evidence Inference tasks following \newcite{paranjape-etal-2020-information} and 2000 training examples for tasks with bigger datasets, viz., MultiRC and FEVER.  We see an improvement of 3.2\% TF1 on BoolQ, 1.7\% on Evidence Inference, 2.9\% in MultiRC and 1.1\% on FEVER. This is desirable since obtaining labeled rationale annotations is expensive. We do not observe any performance improvement for Movie Reviews with IFT. While IFT on NQ or HotpotQA alone improves performance, we find that combining the datasets yields best results. Lastly, while \newcite{narang2020wt5} find that their WT5 model yields a TF1 of 0 on MultiRC when trained on less than 10,000 examples, \ours\ is able to obtain with just 2,000 examples, a TF1 of 65.9\% owing to the use of sentence markers, which improves to 68.8\% with IFT.

\paragraph{Comparison with Prior Work} In Table \ref{tab:comparison} we compare the performance of our best model on the full training set for each dataset, with prior work: 1) The Bert-to-Bert (B2B) supervised model of \newcite{deyoung-etal-2020-eraser}, 2) The Information Bottleneck (IB) approach of \newcite{paranjape-etal-2020-information} and 3) The base version of the seq2seq model of \newcite{narang2020wt5}. The prior best performance for ERASER datasets is distributed among B2B, WT5, and IB. Although \newcite{paranjape-etal-2020-information} only report supervised results using 25\% training data, their model achieves similar performance even with the full training data.

Overall, we outperform prior work in terms of explanation metrics (using TF1 here) on BoolQ (+37.9\% from IB), Movies (+33.7\% from WT5), MultiRC (+2.4\% from WT5), FEVER (+2.6\% from B2B), and EVI (+3.7\% from B2B). We also improve upon Task Accuracy on BoolQ (+12.9\% from IB), EVI (+4.5\% from B2B), MultiRC (+0.9\% from WT5), and FEVER (+4.8\% from B2B). In summary, \ours\ significantly improves the state-of-the-art on multiple ERASER datasets, in both fully supervised and few-shot settings, with each component from Section \ref{sec:model} individually contributing to overall performance.

 We conduct an error analysis on predictions from our best \ours\ model on 50 randomly chosen examples from the validation set of BoolQ, that have a non-perfect rationale F1 score. (see Table \ref{tab:ea}). The two largest sources of errors are: 1) \textit{Overlap and Adequate} - For 36\% of cases, the set of predicted explanations is adequate by itself and overlaps with the true explanations; i.e. the true explanation set contains redundancies, and 2) \textit{Over-prediction} - For 30\% of cases, the set of model predictions are a strict superset of the true explanations. Other sources of errors are; \textit{Overlap and Inadequate} - when the predictions are inadequate but overlap with the true explanations, \textit{No-overlap and Adequate/Inadequate} - when the predictions have no overlap with the true explanations and are either still adequate or inadequate. Since ERASER provides only one set of the multiple possible explanation sets, 8\% non-overlapping predictions also happen to be adequate. \textit{Prediction not in input} - when sentence markers that do not exist in the input are predicted, and \textit{Input Truncated} - when the true explanation sentences are not provided to the model owing to input truncation, which still happens for very long inputs even with a context size of 10. We present examples of these error cases  from the validation set of BoolQ in Appendix \ref{sec:appendix}, and highlight the gold and predicted rationales, together with the overlap between them.

\begin{table}
  \small
  \centering
  \begin{tabular}{lr}
    \toprule
     \textbf{Error Type}  & \textbf{\% Cases} \\
    \midrule
    Overlap and Adequate & $36$ \\
    Overlap and Inadequate & $4$ \\
    Over-Prediction & $30$ \\
    No-overlap and Inadequate & $12$ \\
    No-overlap and Adequate & $8$ \\
    Prediction not in input & $4$ \\
    Input Truncated & $6$ \\
    \bottomrule
  \end{tabular}
  \caption{Frequency distribution of error types in 50 randomly sampled examples with a non-perfect RF1 score, from the validation set of BoolQ, using our best \ours\ model.}
  \label{tab:ea}
\end{table}

\section{Conclusion}

In this paper, we develop generally applicable methods to improve the performance of large pre-trained seq2seq models for jointly producing NL rationales together with answer predictions. Specifically, we introduce sentence markers into seq2seq models to tackle explanation fabrication, we train with larger input passage sizes using the Fusion-in-Decoder architecture, and we infuse knowledge by fine-tuning on re-structured QA datasets. Our methods improve the state of the art on rationale extraction metrics and task accuracy on a number of ERASER benchmarks while reducing the extent to which seq2seq models fabricate explanations to justify incorrect predictions, thereby improving the reliability and verifiability of the generated rationales.

\bibliography{custom}
\bibliographystyle{acl_natbib}

\clearpage
\appendix
\onecolumn
\section{Error Analysis on BoolQ}
\label{sec:appendix}
In Section \ref{sec:results}, we discussed the results of our error analysis of model predictions using 50 randomly chosen examples from the validation set of BoolQ. Here, we present examples for each error type to illustrate the cases with a  non-perfect rationale F1 score. We have preserved the sentence markers (SM) in the document to help locate the gold and predicted sentences easily. The error types are:
\begin{enumerate}
  \item \textit{Overlap and Adequate} - Predicted explanations are adequate and overlap with the true explanations
  \item \textit{Overlap and Inadequate} - Predicted explanations are inadequate but overlap with the true explanations
  \item \textit{Over-prediction} - Predicted explanations are a strict superset of the true explanations
  \item \textit{No overlap and Inadequate} - Predicted explanations are inadequate and do not overlap with the true explanations
  \item \textit{No overlap and Adequate} - Predicted explanations are adequate but do not overlap with the true explanations
  \item \textit{Prediction not in input} - Predicted explanation sentence markers are not in the input 
  \item \textit{Input Truncated} - True explanation sentence markers are not in input 
\end{enumerate}

\begin{table*}[ht]
    \centering
    \begin{tabular}{p{15.5cm}}
    \caption*{\emph{Legend}: \textbf{Sentence Marker (SM)}, \textcolor{green}{Correctly Predicted SM}, \textcolor{red}{Missed SM}, \textcolor{blue}{Over-predicted SM}} \\
    \toprule
    $Overlap\ and\ Adequate$ \\
    \midrule
    Question: {\fontfamily{cmr}\selectfont are the oakland raiders in las vegas now} \\
    Gold Answer: {\fontfamily{cmr}\selectfont True} \\
    Predicted Answer: {\fontfamily{cmr}\selectfont True} \\
    Gold Rationales: {\fontfamily{cmr}\selectfont ['S0', 'S1', 'S2', 'S3', 'S4', 'S5']} \\
    Predicted Rationales': {\fontfamily{cmr}\selectfont ['S0', 'S1', 'S2', 'S3', 'S4']} \\
    Document: {\fontfamily{cmr}\selectfont \textcolor{green}{\textbf{S0}} OAKLAND RAIDERS RELOCATION TO LAS VEGAS \textcolor{green}{\textbf{S1}} The Oakland Raiders relocation to Las Vegas is a successful effort by the owner of the Oakland Raiders ( Mark Davis ) to relocate the American football club from its current and longtime home of Oakland , California to Paradise , Nevada . \textcolor{green}{\textbf{S2}} The team is scheduled to begin playing its home games at the Las Vegas Stadium as the Las Vegas Raiders for the 2020 National Football League ( NFL ) season , although the Raiders could move to and begin playing home games at Sam Boyd Stadium in Whitney , Nevada for the 2019 season . \textcolor{green}{\textbf{S3}} NFL team owners voted 31–1 to approve the move , which was announced at the annual league meetings in Phoenix , Arizona on March 27 , 2017 . \textcolor{green}{\textbf{S4}} The Raiders became the third NFL franchise to relocate in the 2010s , following the Rams \' move from St. Louis , Missouri to Los Angeles , California on January 12 , 2016 , and the Chargers \' move from San Diego , California to Los Angeles on January 12 , 2017 . \textcolor{red}{\textbf{S5}} The Raiders \' move to Las Vegas comes after years of failed efforts to renovate or replace the Oakland – Alameda County Coliseum , which has been rated by multiple sources as one of the worst stadiums in the NFL . \textbf{\dots} \textbf{S113} Davis publicly reiterated his commitment to his announced plans to relocate the Raiders franchise to Las Vegas , Nevada with the support of the state of Nevada and casino mogul Sheldon Adelson , and said he did not wish to negotiate further with Oakland while the Las Vegas deal was still actively in progress ; any relocation to Las Vegas needed to be approved by a three - quarters majority of all NFL owners , and NFL commissioner Roger Goodell publicly stated his preference for keeping the Raiders franchise in Oakland if at all possible .} \\
    \midrule
    Question: {\fontfamily{cmr}\selectfont is a woodchuck and a groundhog the same} \\
    Gold Answer: {\fontfamily{cmr}\selectfont True} \\
    Predicted Answer: {\fontfamily{cmr}\selectfont True} \\
    Gold Rationales: {\fontfamily{cmr}\selectfont ['S1', 'S2', 'S3', 'S4', 'S5', 'S6', 'S7', 'S8', 'S9']} \\
    Predicted Rationales': {\fontfamily{cmr}\selectfont ['S0', 'S1', 'S2', 'S3']} \\
    Document: {\fontfamily{cmr}\selectfont \textcolor{blue}{\textbf{S0}} GROUNDHOG \textcolor{green}{\textbf{S1}} The groundhog ( Marmota monax ) , also known as a woodchuck , is a rodent of the family Sciuridae , belonging to the group of large ground squirrels known as marmots . \textcolor{green}{\textbf{S2}} It was first scientifically described by Carl Linnaeus in 1758 . \textcolor{green}{\textbf{S3}} The groundhog is also referred to as a chuck , wood - shock , groundpig , whistlepig , whistler , thickwood badger , Canada marmot , monax , moonack , weenusk , red monk and , among French Canadians in eastern Canada , siffleux . \textcolor{red}{\textbf{S4}} The name " thickwood badger " was given in the Northwest to distinguish the animal from the prairie badger . \textcolor{red}{\textbf{S5}} Monax ( Móonack ) is an Algonquian name of the woodchuck , which meant " digger " ( cf . \textcolor{red}{\textbf{S6}} Lenape monachgeu ) . \textcolor{red}{\textbf{S7}} Young groundhogs may be called chucklings . \textcolor{red}{\textbf{S8}} Other marmots , such as the yellow - bellied and hoary marmots , live in rocky and mountainous areas , but the groundhog is a lowland creature . \textcolor{red}{\textbf{S9}} It is found through much of the eastern United States across Canada and into Alaska DESCRIPTION Section::::Description . \textbf{\dots} \textbf{S159} * Woodchuck ( Groundhog ) , Missouri Conservation Commission * Breeding and Experimental Facility for Woodchucks} \\
    \bottomrule
    \end{tabular}
\end{table*}
\begin{table*}[ht]
    \centering
    \begin{tabular}{p{15.5cm}}
    \caption*{\emph{Legend}: \textbf{Sentence Marker (SM)}, \textcolor{green}{Correctly Predicted SM}, \textcolor{red}{Missed SM}, \textcolor{blue}{Over-predicted SM}} \\
    \toprule
    $Overlap\ and\ Inadequate$ \\
    \midrule
    Question: {\fontfamily{cmr}\selectfont are all mass air flow sensors the same} \\
    Gold Answer: {\fontfamily{cmr}\selectfont False} \\
    Predicted Answer: {\fontfamily{cmr}\selectfont False} \\
    Gold Rationales: {\fontfamily{cmr}\selectfont ['S4', 'S5', 'S6', 'S7', 'S8']} \\
    Predicted Rationales': {\fontfamily{cmr}\selectfont ['S0', 'S1', 'S2', 'S3', 'S4']} \\
    Document: {\fontfamily{cmr}\selectfont {'\textcolor{blue}{\textbf{S0}} MASS FLOW SENSOR \textcolor{blue}{\textbf{S1}} A mass ( air ) flow sensor ( MAF ) is a sensor used to determine the mass flow rate of air entering a fuel - injected internal combustion engine . \textcolor{blue}{\textbf{S2}} The air mass information is necessary for the engine control unit ( ECU ) to balance and deliver the correct fuel mass to the engine . \textcolor{blue}{\textbf{S3}} Air changes its density with temperature and pressure . \textcolor{green}{\textbf{S4}} In automotive applications , air density varies with the ambient temperature , altitude and the use of forced induction , which means that mass flow sensors are more appropriate than volumetric flow sensors for determining the quantity of intake air in each cylinder . \textcolor{red}{\textbf{S5}} There are two common types of mass airflow sensors in use on automotive engines . \textcolor{red}{\textbf{S6}} These are the vane meter and the hot wire . \textcolor{red}{\textbf{S7}} Neither design employs technology that measures air mass directly . \textcolor{red}{\textbf{S8}} However , with additional sensors and inputs , an engine \'s ECU can determine the mass flow rate of intake air . \textbf{\dots} \textbf{S103} REFERENCES EXTERNAL LINKS * A Hot Film sensor with theory of operation * A video example of cleaning a MAF sensor * An example of how to clean a MAF sensor , \textbf{S104} 3 wire \textbf{S105} * How To Test a MAF}} \\
    \bottomrule
    \end{tabular}
\end{table*}
\begin{table*}[ht]
    \centering
    \begin{tabular}{p{15.5cm}}
    \caption*{\emph{Legend}: \textbf{Sentence Marker (SM)}, \textcolor{green}{Correctly Predicted SM}, \textcolor{red}{Missed SM}, \textcolor{blue}{Over-predicted SM}} \\
    \toprule
    $Over-Prediction$ \\
    \midrule
    Question: {\fontfamily{cmr}\selectfont does the tour de france take the same route every year} \\
    Gold Answer: {\fontfamily{cmr}\selectfont False} \\
    Predicted Answer: {\fontfamily{cmr}\selectfont False} \\
    Gold Rationales: {\fontfamily{cmr}\selectfont ['S8', 'S9', 'S10', 'S11', 'S12']} \\
    Predicted Rationales': {\fontfamily{cmr}\selectfont ['S8', 'S9', 'S10', 'S11', 'S12', 'S13', 'S14', 'S15', 'S16', 'S17']} \\
    Document: {\fontfamily{cmr}\selectfont {\textbf{S0} TOUR DE \textbf{S1} FRANCE \textbf{\dots} \textcolor{green}{\textbf{S8}} The Tour is a UCI World Tour event , which means that the teams that compete in the race are mostly UCI WorldTeams , with the exception of the teams that the organizers invite . \textcolor{green}{\textbf{S9}} Traditionally , the race is held primarily in the month of July . \textcolor{green}{\textbf{S10}} While the route changes each year , the format of the race stays the same with the appearance of time trials , the passage through the mountain chains of the Pyrenees and the Alps , and the finish on the Champs - Élysées in Paris . \textcolor{green}{\textbf{S11}} The modern editions of the Tour de France consist of 21 day - long segments ( stages ) over a 23-day period and cover around . \textcolor{green}{\textbf{S12}} The race alternates between clockwise and counterclockwise circuits of France . \textcolor{blue}{\textbf{S13}} There are usually between 20 and 22 teams , with eight riders in each . \textcolor{blue}{\textbf{S14}} All of the stages are timed to the finish ; the riders ' times are compounded with their previous stage times . \textcolor{blue}{\textbf{S15}} The rider with the lowest cumulative finishing times is the leader of the race and wears the yellow jersey . explain boolq question: does the tour de france take the same route every year passage: \textcolor{blue}{\textbf{S16}} While the general classification garners the most attention , there are other contests held within the Tour : the points classification for the sprinters , the mountains classification for the climbers , young rider classification for riders under the age of 26 , and the team classification for the fastest teams . \textcolor{blue}{\textbf{S17}} Achieving a stage win also provides prestige , often accomplished by a team 's cycling sprinter specialist . \textbf{\dots}} \textbf{S161} He helped drive an internationalization of the Tour de France , and cycling in general .} \\
    \midrule
    Question: {\fontfamily{cmr}\selectfont was kentucky a southern state in the civil war} \\
    Gold Answer: {\fontfamily{cmr}\selectfont False} \\
    Predicted Answer: {\fontfamily{cmr}\selectfont False} \\
    Gold Rationales: {\fontfamily{cmr}\selectfont ['S4', 'S5', 'S6', 'S7']} \\
    Predicted Rationales': {\fontfamily{cmr}\selectfont ['S0', 'S1', 'S2', 'S3', 'S4', 'S5', 'S6']} \\
    Document: {\fontfamily{cmr}\selectfont {\textcolor{blue}{\textbf{S0}} KENTUCKY IN THE AMERICAN CIVIL WAR Kentucky was a border state of key importance in the American Civil War . \textcolor{blue}{\textbf{S1}} President Abraham Lincoln recognized the importance of the Commonwealth when , in a September 1861 letter to Orville Browning , he wrote : I think to lose Kentucky is nearly the same as to lose the whole game . \textcolor{blue}{\textbf{S2}} Kentucky gone , we can not hold Missouri , nor Maryland . \textcolor{blue}{\textbf{S3}} These all against us , and the job on our hands is too large for us . \textcolor{green}{\textbf{S4}} We would as well consent to separation at once , including the surrender of this capitol . \textcolor{green}{\textbf{S5}} Kentucky , being a border state , was among the chief places where the " Brother against brother " scenario was prevalent . \textcolor{green}{\textbf{S6}} Kentucky officially declared its neutrality at the beginning of the war , but after a failed attempt by Confederate General Leonidas Polk to take the state of Kentucky for the Confederacy , the legislature petitioned the Union Army for assistance . \textcolor{red}{\textbf{S7}} After early 1862 Kentucky came largely under Union control . \textbf{\dots} \textbf{S128} Union ironclads routed the Confederate river gunboats on the Mississippi River during the Battle of Lucas Bend on January 11 , forcing them back to Columbus .}} \\
    \bottomrule
    \end{tabular}
\end{table*}
\begin{table*}[ht]
    \centering
    \begin{tabular}{p{15.5cm}}
    \caption*{\emph{Legend}: \textbf{Sentence Marker (SM)}, \textcolor{green}{Correctly Predicted SM}, \textcolor{red}{Missed SM}, \textcolor{blue}{Over-predicted SM}} \\
    \toprule
    $No\ overlap\ and\ Inadequate$ \\
    \midrule
    Question: {\fontfamily{cmr}\selectfont are there mountains in the state of indiana} \\
    Gold Answer: {\fontfamily{cmr}\selectfont False} \\
    Predicted Answer: {\fontfamily{cmr}\selectfont True} \\
    Gold Rationales: {\fontfamily{cmr}\selectfont ['S107', 'S108']} \\
    Predicted Rationales': {\fontfamily{cmr}\selectfont ['S84', 'S85', 'S86']} \\
    Document: {\fontfamily{cmr}\selectfont {\textbf{S0} GEOGRAPHY OF INDIANA \textbf{S1} The geography of Indiana comprises the physical features of the land and relative location of U.S. State of Indiana . \textbf{\dots} \textcolor{blue}{\textbf{S84}} Rural areas in the central portion of the state are typically composed of a patchwork of fields and forested areas . \textcolor{blue}{\textbf{S85}} The geography of Central Indiana consists of gently rolling hills and sandstone ravines carved out by the retreating glaciers . explain boolq question: are there mountains in the state of indiana passage: \textcolor{blue}{\textbf{S86}} Many of these ravines can be found in west - central Indiana , specifically along Sugar Creek in Turkey Run State Park and Shades State Park . \textbf{\dots} \textcolor{red}{\textbf{S107}} PHYSIOGRAPHY Section::::Physiography . \textcolor{red}{\textbf{S108}} Indiana is broken up into three main physical regions : The Great Lakes Plain in the northern third of the state , the Tipton Till Plain in the central third , and the Southern Hills and Lowlands region in the southern third . \textbf{\dots} \textbf{S136} * Midwestern United States NOTES REFERENCES}} \\
    \midrule
    Question: {\fontfamily{cmr}\selectfont can a company have a social security number} \\
    Gold Answer: {\fontfamily{cmr}\selectfont False} \\
    Predicted Answer: {\fontfamily{cmr}\selectfont True} \\
    Gold Rationales: {\fontfamily{cmr}\selectfont ['S0', 'S1', 'S2', 'S3']} \\
    Predicted Rationales': {\fontfamily{cmr}\selectfont ['S28', 'S29', 'S30', 'S31']} \\
    Document: {\fontfamily{cmr}\selectfont {\textcolor{red}{\textbf{S0}} SOCIAL SECURITY NUMBER \textcolor{red}{\textbf{S1}} In the United States , a Social Security number ( SSN ) is a nine - digit number issued to U.S. citizens , permanent residents , and temporary ( working ) residents under section 205(c)(2 ) of the Social Security Act , codified as . \textcolor{red}{\textbf{S2}} The number is issued to an individual by the Social Security Administration , an independent agency of the United States government . \textcolor{red}{\textbf{S3}} Although its primary purpose is to track individuals for Social Security purposes , the Social Security number has become a de facto national identification number for taxation and other purposes . \textbf{\dots} \textcolor{blue}{\textbf{S28}} NON - UNIVERSAL STATUS Section::::Non - universal status . \textcolor{blue}{\textbf{S29}} Social Security was originally a universal tax , but when Medicare was passed in 1965 , objecting religious groups in existence prior to 1951 were allowed to opt out of the system . \textcolor{blue}{\textbf{S30}} Because of this , not every American is part of the Social Security program , and not everyone has a number . \textcolor{blue}{\textbf{S31}} However , a social security number is required for parents to claim their children as dependents for federal income tax purposes , and the Internal Revenue Service requires all corporations to obtain SSNs ( or alternative identifying numbers ) from their employees , as described below . \textbf{\dots} \textbf{S150} LIST OF SOCIAL SECURITY AREA NUMBERS Section::::List of Social Security Area Numbers .}} \\
    \bottomrule
    \end{tabular}
\end{table*}
\begin{table*}[ht]
    \centering
    \begin{tabular}{p{15.5cm}}
    \caption*{\emph{Legend}: \textbf{Sentence Marker (SM)}, \textcolor{green}{Correctly Predicted SM}, \textcolor{red}{Missed SM}, \textcolor{blue}{Over-predicted SM}} \\
    \toprule
    $No\ overlap\ and\ Adequate$ \\
    \midrule
    Question: {\fontfamily{cmr}\selectfont can you make and receive calls in airplane mode} \\
    Gold Answer: {\fontfamily{cmr}\selectfont False} \\
    Predicted Answer: {\fontfamily{cmr}\selectfont True} \\
    Gold Rationales: {\fontfamily{cmr}\selectfont ['S0', 'S1', 'S2']} \\
    Predicted Rationales': {\fontfamily{cmr}\selectfont ['S10', 'S11', 'S12', 'S13']} \\
    Document: {\fontfamily{cmr}\selectfont {\textcolor{red}{\textbf{S0}} AIRPLANE MODE Airplane mode , aeroplane mode , flight mode , offline mode , or standalone mode is a setting available on smartphones and other portable computers that , when activated , suspends radio - frequency signal transmission by the device , thereby disabling Bluetooth , telephony , and Wi - Fi . \textcolor{red}{\textbf{S1}} GPS may or may not be disabled , because it does not involve transmitting radio waves . \textcolor{red}{\textbf{S2}} The name comes from the prohibition by most of the airlines of using equipment transmitting radio - frequency signal while in flight ; using airplane mode prevents devices from transmitting . \textbf{\dots} \textcolor{blue}{\textbf{S10}} The statement cites the common practice of aircraft operators whose aircraft can tolerate use of these personal electronic devices , but use may still be prohibited on some models of aircraft . \textcolor{blue}{\textbf{S11}} While in airplane mode , most devices allow the user to continue to use their email client or other program to write text or E - mail messages which are saved in memory to send when airplane mode is disabled . \textcolor{blue}{\textbf{S12}} Although it is not possible to make normal calls or send text in airplane mode , devices such as some Nokia smartphones allow the user to make calls to emergency services even in airplane mode , while others do not . explain boolq question: can you make and receive calls in airplane mode passage: \textcolor{blue}{\textbf{S13}} As a side - effect , airplane mode reduces power consumption and increases battery endurance by shutting down the device 's transmitters and receivers . \textbf{\dots} \textbf{S20} REFERENCES EXTERNAL LINKS \textbf{S21} * Copa Airlines ' cell phone policy \textbf{S22} * QANTAS policy on usage of flight mode}} \\
    \midrule
    Question: {\fontfamily{cmr}\selectfont is row row row your boat a masonic poem} \\
    Gold Answer: {\fontfamily{cmr}\selectfont False} \\
    Predicted Answer: {\fontfamily{cmr}\selectfont False} \\
    Gold Rationales: {\fontfamily{cmr}\selectfont ['S0', 'S1', 'S2', 'S3']} \\
    Predicted Rationales': {\fontfamily{cmr}\selectfont ['S33', 'S34', 'S35', 'S36', 'S37']} \\
    Document: {\fontfamily{cmr}\selectfont {\textcolor{red}{\textbf{S0}} ROW , ROW , ROW \textcolor{red}{\textbf{S1}} YOUR BOAT " Row , Row , Row Your Boat " is an English language nursery rhyme and a popular children 's song . \textcolor{red}{\textbf{S2}} It can also be an " action " nursery rhyme , whose singers sit opposite one another and " row " forwards and backwards with joined hands . \textcolor{red}{\textbf{S3}} It has a Roud Folk Song Index number of 19236 . \textbf{\dots} \textcolor{blue}{\textbf{S33}} ly , ORIGINS Section::::Origins . \textcolor{blue}{\textbf{S34}} It has been suggested that the song may have originally arisen out of American minstrelsy . \textcolor{blue}{\textbf{S35}} The earliest printing of the song is from 1852 , when the lyrics were published with similar lyrics to those used today , but with a very different tune . \textcolor{blue}{\textbf{S36}} It was reprinted again two years later with the same lyrics and another tune . \textcolor{blue}{\textbf{S37}} The modern tune was first recorded with the lyrics in 1881 , mentioning Eliphalet Oram Lyte in The Franklin Square Song Collection but not making it clear whether he was the composer or adapter . \textbf{\dots} \textbf{S42} Don Music , a muppet character in Sesame Street , changed the lyrics to feature a car instead of a boat . \textbf{S43} Versions include : And : NOTES AND REFERENCES}} \\
    \bottomrule
    \end{tabular}
\end{table*}
\begin{table*}[ht]
    \centering
    \begin{tabular}{p{15.5cm}}
    \caption*{\emph{Legend}: \textbf{Sentence Marker (SM)}, \textcolor{green}{Correctly Predicted SM}, \textcolor{red}{Missed SM}, \textcolor{blue}{Over-predicted SM}} \\
    \toprule
    $Prediction\ not\ in\ Input$ \\
    \midrule
    Question: {\fontfamily{cmr}\selectfont is there taco bell on the east coast} \\
    Gold Answer: {\fontfamily{cmr}\selectfont True} \\
    Predicted Answer: {\fontfamily{cmr}\selectfont False} \\
    Gold Rationales: {\fontfamily{cmr}\selectfont ['S52', 'S53', 'S54', 'S55', 'S56', 'S57', 'S58', 'S59', 'S60', 'S61', 'S62']} \\
    Predicted Rationales': {\fontfamily{cmr}\selectfont ['S261', 'S262', 'S263', 'S264', 'S265']} \\
    Document: {\fontfamily{cmr}\selectfont {\textbf{S0} TACO \textbf{S1} BELL Taco Bell is an American chain of fast food restaurants based out of Irvine , California and a subsidiary of Yum ! \textbf{\dots} \textcolor{red}{\textbf{S52}} Brands on May 16 , 2002 . \textcolor{red}{\textbf{S53}} Taco Bell began experimenting with fast - casual and urban concepts when it created U.S. Taco Co. and Urban Taproom in 2014 . \textcolor{red}{\textbf{S54}} The menu consists of tacos with American fillings , and did not sell food sold in Taco Bell restaurants such as burritos . \textcolor{red}{\textbf{S55}} It was launched in Huntington Beach , California in August 2014 . \textcolor{red}{\textbf{S56}} U.S. Taco Co. closed on September 15 , 2015 \textcolor{red}{\textbf{S57}} so the company could focus on its new similar Taco Bell Cantina concept , which featured special menu items and served alcohol . \textcolor{red}{\textbf{S58}} It opened its first location a few days later in Chicago 's Wicker Park neighborhood , followed by a location in San Francisco about a month later , located less than a block away from AT\&T Park . \textcolor{red}{\textbf{S59}} In 2016 , Taco Bell launched the Taco Bell Catina flagship store located on the Las Vegas strip . \textcolor{red}{\textbf{S60}} This 24-hour restaurant hosts a variety of unique features including alcohol , new menu items , and a DJ . \textcolor{red}{\textbf{S61}} It was announced in August 2017 that the store would begin hosting weddings . explain boolq question: is there taco bell on the east coast passage: \textcolor{red}{\textbf{S62}} Taco Bell Cantina currently has locations in San Francisco , Berkeley , Chicago ( 2 locations ) , Las Vegas , Austin , Fayetteville , Cincinnati , Cleveland , Atlanta , Newport Beach , and plans to open soon in Somerville , MA . \textbf{\dots} \textbf{S137} In April 2017 , Taco Bell announced that it will begin testing the Naked Breakfast Taco in Flint , Michigan in mid - April .}} \\
    \midrule
    Question: {\fontfamily{cmr}\selectfont is costa rica part of the ring of fire} \\
    Gold Answer: {\fontfamily{cmr}\selectfont True} \\
    Predicted Answer: {\fontfamily{cmr}\selectfont True} \\
    Gold Rationales: {\fontfamily{cmr}\selectfont ['S121', 'S122', 'S123', 'S124', 'S125', 'S126', 'S127', 'S128']} \\
    Predicted Rationales': {\fontfamily{cmr}\selectfont ['S261', 'S262', 'S263', 'S264', 'S265', 'S266', 'S267', 'S268', 'S269', 'S270']} \\
    Document: {\fontfamily{cmr}\selectfont {\textbf{S0} RING OF FIRE \textbf{S1} The Ring of Fire is a major area in the basin of the Pacific Ocean where many earthquakes and volcanic eruptions occur . \textbf{\dots} \textcolor{red}{\textbf{S121}} AMERICA COSTA RICA Section::::Central America . \textcolor{red}{\textbf{S122}} Section::::Costa Rica . \textcolor{red}{\textbf{S123}} The Volcanological and Seismological Observatory of Costa Rica ( OVSICORI ) at the National University of Costa Rica , in Spanish Observatorio Vulcanológico y Sismológico de Costa Rica ( OVSICORI ) have a dedicated team in charge of researching and monitoring the volcanoes , earthquakes , and other tectonic processes in the Central America Volcanic Arc . explain boolq question: is costa rica part of the ring of fire passage: \textcolor{red}{\textbf{S124}} In 1984 , the OVSICORI - A initiated the operation of a seismographic network designed to monitor seismic and volcanic activity throughout the national territory . \textcolor{red}{\textbf{S125}} Currently , the seismographic network has an analog and a digital registration system . \textcolor{red}{\textbf{S126}} The latter enables online analysis of seismic signals , allowing to expedite the analysis of signals and the study using modern computerized methods . \textcolor{red}{\textbf{S127}} Poás Volcano is an active stratovolcano located in central Costa Rica ; it has erupted 39 times since 1828 . \textcolor{red}{\textbf{S128}} On February 25 , 2014 , a webcam from the OVSICORI captured the moment a dark cloud exploded about in the air from a massive crater of the Poás Volcano . \textbf{\dots} \textbf{S138} A few other active volcanoes in northern Mexico are related to extensional tectonics of the Basin and Range Province , which splits the Baja California peninsula from the mainland .}} \\
    \bottomrule
    \end{tabular}
\end{table*}
\begin{table*}[ht]
    \centering
    \begin{tabular}{p{15.5cm}}
    \caption*{\emph{Legend}: \textbf{Sentence Marker (SM)}, \textcolor{green}{Correctly Predicted SM}, \textcolor{red}{Missed SM}, \textcolor{blue}{Over-predicted SM}} \\
    \toprule
    $Input\ Truncated$ \\
    \midrule
    Question: {\fontfamily{cmr}\selectfont is there a new series of sherlock holmes} \\
    Gold Answer: {\fontfamily{cmr}\selectfont False} \\
    Predicted Answer: {\fontfamily{cmr}\selectfont False} \\
    Gold Rationales: {\fontfamily{cmr}\selectfont ['S239', 'S240', 'S241']} \\
    Predicted Rationales': {\fontfamily{cmr}\selectfont ['S0', 'S1', 'S2', 'S3', 'S4', 'S5', 'S6']} \\
    Document: {\fontfamily{cmr}\selectfont {\textcolor{blue}{\textbf{S0}} SHERLOCK ( TV SERIES ) \textcolor{blue}{\textbf{S1}} Sherlock is a British crime drama television series based on Sir Arthur Conan Doyle 's Sherlock Holmes detective stories . \textcolor{blue}{\textbf{S2}} Created by Steven Moffat and Mark Gatiss , it stars Benedict Cumberbatch as Sherlock Holmes and Martin Freeman as Doctor John Watson . \textcolor{blue}{\textbf{S3}} Thirteen episodes have been produced , with four three - part series airing from 2010 to 2017 , and a special episode that aired on 1 January 2016 . \textcolor{blue}{\textbf{S4}} The series is set in the present day , while the one - off special features a Victorian period fantasy resembling the original Holmes stories . \textcolor{blue}{\textbf{S5}} Sherlock is produced by the British network BBC , along with Hartswood Films , with Moffat , Gatiss , Sue Vertue and Rebecca Eaton serving as executive producers . \textcolor{blue}{\textbf{S6}} The series is supported by the American station WGBH Boston for its Masterpiece anthology series on PBS , where it also airs in the United States . \textbf{\dots} \textbf{S142} Arnold explains that he and Price worked with the producers to " come up with a central theme and character " for the series , then found what was " going to be the defining sound of this show " .}} \\
    \midrule
    Question: {\fontfamily{cmr}\selectfont did the harry potter movies win any oscars} \\
    Gold Answer: {\fontfamily{cmr}\selectfont False} \\
    Predicted Answer: {\fontfamily{cmr}\selectfont True} \\
    Gold Rationales: {\fontfamily{cmr}\selectfont ['S297', 'S298', 'S299', 'S300']} \\
    Predicted Rationales': {\fontfamily{cmr}\selectfont ['S261', 'S262', 'S263', 'S264', 'S265']} \\
    Document: {\fontfamily{cmr}\selectfont {\textbf{S0} HARRY POTTER ( FILM SERIES ) \textbf{S1} Harry Potter is a British - American film series based on the Harry Potter novels by author J. K. Rowling . \textbf{S2} The series is distributed by Warner Bros. and consists of eight fantasy films , beginning with Harry Potter and the Philosopher 's Stone ( 2001 ) and culminating with Harry Potter and the Deathly Hallows – Part 2 ( 2011 ) . \textbf{\dots} \textbf{S125} CAST AND CREW Section::::Cast and crew . \textbf{S126} Aside from the three lead actors , other notable cast members include Robbie Coltrane as Rubeus Hagrid , Tom Felton as Draco Malfoy , Alan Rickman as Severus Snape , and Dame Maggie Smith as Minerva McGonagall .}} \\
    \bottomrule
    \end{tabular}
\end{table*}
\end{document}